\pdfoutput=1

\documentclass[11pt]{article}

\usepackage{ACL2023}

\usepackage{times}
\usepackage{latexsym}

\usepackage[T1]{fontenc}

\usepackage[utf8]{inputenc}

\usepackage{microtype}

\usepackage{inconsolata}

\usepackage{enumitem}
\usepackage{array}
\usepackage{graphicx}
\usepackage{multirow}
\usepackage{amsmath}
\usepackage{listings}
\usepackage{float}
\usepackage{subcaption}
\usepackage{cleveref}
\crefformat{footnote}{#2\footnotemark[#1]#3}
\usepackage{placeins}

\usepackage{pifont}
\newcommand{\cmark}{\text{\ding{51}}}%
\newcommand{\xmark}{\text{\ding{55}}}%

\newcommand{\aff}[1]{\textsuperscript{#1}}
\title{LLMs cannot \textit{find} reasoning errors, but can \textit{correct} them given the error location}

\author{Gladys Tyen*\aff{1}, Hassan Mansoor\aff{2}, Victor Cărbune\aff{2}, Peter Chen\textdagger\aff{2}, Tony Mak\textdagger\aff{2} \\
  \aff{1}University of Cambridge, Dept. of Computer Science \& Technology, ALTA Institute \\
  \aff{2}Google Research \\
  \texttt{gladys.tyen@cl.cam.ac.uk} \\
  \texttt{\{vcarbune,hassan,chenfeif,tonymak\}@google.com}}

\begin{document}
\maketitle
\begingroup\def\thefootnote{*}\footnotetext{Work done during an internship at Google Research.}\endgroup
\begingroup\def\thefootnote{\textdagger}\footnotetext{Equal leadership contribution.}\endgroup
\begin{abstract}
While self-correction has shown promise in improving LLM outputs in terms of style and quality \citep[e.g.][]{chen2023iterative,madaan2023self}, recent attempts to self-correct logical or reasoning errors often cause correct answers to become incorrect, resulting in worse performances overall \citep{huang2023large}. 
In this paper, we show that poor self-correction performance stems from LLMs' inability to \textit{find} logical mistakes, rather than their ability to \textit{correct} a known mistake. Firstly, we benchmark several state-of-the-art LLMs on their mistake-finding ability and demonstrate that they generally struggle with the task, even in highly objective, unambiguous cases. Secondly, we test the \textit{correction} abilities of LLMs -- separately from mistake finding -- using a backtracking setup that feeds ground truth mistake location information to the model. We show that this boosts downstream task performance across our 5 reasoning tasks, indicating that LLMs' \textit{correction} abilities are robust. Finally, we show that it is possible to obtain mistake location information \textit{without} ground truth labels or in-domain training data. We train a small classifier with out-of-domain data, which exhibits stronger mistake-finding performance than prompting a large model. We release our dataset of LLM-generated logical mistakes, BIG-Bench Mistake, to enable further research into locating LLM reasoning mistakes.
\end{abstract}

\section{Introduction}
Large Language Models (LLMs) have dominated the field of NLP in recent years, achieving state-of-the-art performance in a large variety of applications. In particular, LLMs have demonstrated the ability to solve tasks with zero- or few-shot prompting, giving rise to prompting methods such as Chain-of-Thought (CoT) \citep{wei2022chain}, Self-Consistency (SC) \citep{wang2023selfconsistency}, ReAct \citep{yao2022react}, etc.

Recent literature on few- or zero-shot prompting has focused on the concept of \textit{self-correction}, i.e. having an LLM correct its own outputs \citep{shinn2023reflexion,miao2024selfcheck,madaan2023self,chen2023iterative,saunders2022self}. (See \citet{pan2023automatically} for a review of the literature.)

However, \citet{huang2023large} note that while self-correction may prove effective for improving model outputs in terms of style and quality, when it comes to \textit{reasoning} tasks, LLMs struggle to identify and fix errors without external feedback: for example, Reflexion \citep{shinn2023reflexion} and RCI \citep{kim2023language} both use ground truth correctness as a signal to halt the self-correction loop. Initially observed by \citet{madaan2023self} on a math dataset, \citet{huang2023large} further demonstrate this shortcoming of self-correction in 2 additional datasets.

While previous work typically present self-correction as a single process, we divide it into \textbf{mistake finding} and \textbf{output correction} to better understand each component individually.

\textbf{Mistake finding} is a fundamental reasoning skill that has been studied and utilised extensively in philosophy, psychology, and mathematics, spawning concepts such as critical thinking, and logical and mathematical fallacies. One might expect that the ability to find mistakes should also be an important requirement for LLMs. However, our results show that state-of-the-art LLMs currently \textit{cannot} find mistakes reliably.

\textbf{Output correction} involves partially or completely changing previously generated outputs. With self-correction, this is typically done with outputs generated by the same model (see \citet{pan2023automatically}). Despite LLMs' inability to \textit{find} mistakes, our results show that they can \textit{correct} outputs, if given information about the mistake location. 
While LLMs struggle with mistake-finding in few-shot conditions, we can obtain more reliable mistake location information using a small, trained classifier.

Our contributions for this paper are as follows:
\vspace{-0.2cm}
\begin{enumerate}[topsep=0.2cm,itemsep=0cm,leftmargin=0.5cm]
    \item With Chain-of-Thought prompting, any task can be turned into a mistake-finding task. We collect and release\footnote{Publicly available at \url{https://github.com/WHGTyen/BIG-Bench-Mistake}.} to the research community \textbf{BIG-Bench Mistake}, a dataset of CoT-style traces\footnote{We refer to a set of CoT reasoning steps as a \textit{trace}.} generated using PaLM 2 \citep{anil2023palm}, and annotated according to where the first logical mistake is. To our knowledge, BIG-Bench Mistake is the first dataset of its kind that goes beyond problems in mathematics.
    \item We produce benchmark results for our dataset to test the reasoning capabilities of five state-of-the-art LLMs. We demonstrate that these LLMs \textbf{struggle with mistake finding, even for objective, unambiguous cases}. We hypothesise that this is a main contributing factor to LLMs' inability to self-correct reasoning errors.
    \item We test LLMs' ability to correct reasoning errors \textit{separately} from mistake-finding, by feeding to the model the ground truth (or \textit{oracle}) mistake location information through a backtracking method. We demonstrate that \textbf{LLMs' correction abilities are robust}, effectively correcting outputs that are originally incorrect, with minimal effect on outputs that are originally correct.
    \item We demonstrate that LLMs' low accuracy at few-shot mistake-finding can be improved upon 
    \textbf{without using in-domain training data}. As a proof-of-concept, we train a small mistake-finding classifier using out-of-domain data, which performs better than prompting a large model. We leave the development of more sophisticated methods to future work.
\end{enumerate}

\section{BIG-Bench Mistake}
\label{sec:dataset}
BIG-Bench Mistake contains 2186 sets of CoT-style traces. Each trace is generated with PaLM 2 Unicorn \citep{anil2023palm}, and annotated with the location of the first logical error. \autoref{tab:example} shows an example trace where the mistake location\footnote{As some traces may not contain mistakes, we use the term \textit{mistake location} as a multi-class label that can refer to either the integer $N$ where the $N$\textsuperscript{th} step contains the first mistake, or that there are no mistakes.} is the 4\textsuperscript{th} step. 

\begin{table}[h]
    \small
    \renewcommand{\arraystretch}{2}%
    \begin{tabular}[t]{>{\arraybackslash}m{1.4cm}>{\arraybackslash}m{5.5cm}}
        \multicolumn{2}{>{\arraybackslash}m{7.3cm}}{\textbf{Sort the following words alphabetically: List: hypochlorite ponderosa phone credulity} \vspace{-0.2cm}}\\
        Thought 1: & I should start by looking at the first letter of the words in the list. The first letter: "hypochlorite": "h" (8). "ponderosa": "p" (16). "phone": "p" (16). "credulity": "c" (3).\\
        \hline
        Thought 2: & We now have: (3) "credulity" < (8) "hypochlorite" < (16) ["ponderosa" ? "phone"]. \\
        \hline
        Thought 3: & Now let's sort this subpart ["ponderosa" ? "phone"] by looking at their second letters. The second letter: "ponderosa": "o" (15). "phone": "h" (8). \\
        \hline
        Thought 4: \newline \textbf{(MISTAKE)} & We now have: (8) "phone" < (15) "ponderosa" for the subpart. Hence, we have "credulity" < "phone" < "ponderosa". \\
        \hline
        Thought 5: & I have now sorted all the words. The answer is credulity hypochlorite phone ponderosa \\
    \end{tabular}
    \vspace{-0.2cm}
    \caption{Example of a CoT trace for the word sorting task. There is a mistake in Thought 4: the ordering "credulity" < "phone" < "ponderosa" is missing the word \textit{hypochlorite}.}
    \label{tab:example}
\end{table}

Our traces span across a set of 5 tasks from the BIG-bench dataset \citep{srivastava2023beyond}: word sorting, tracking shuffled objects, logical deduction, multi-step arithmetic, and Dyck languages\footnote{These 5 tasks are selected because 1) \citet{anil2023palm} demonstrate that PaLM 2 performs poorly on these tasks, so it is likely to generate mistakes in CoT traces; 2) mistakes in these tasks are likely to be unambiguous, therefore minimising subjectivity during annotation; and 3) identifying mistakes for these tasks does not require expertise knowledge.}. CoT prompting is used to prompt PaLM 2 to answer questions from each task. As we wanted to separate our CoT traces into distinct steps, we follow \citet{yao2022react} and generate each step separately, using the newline as a stop token.

All traces are generated with temperature $= 0$. The correctness of answers are determined by exact match. Prompts can be found at \url{https://github.com/WHGTyen/BIG-Bench-Mistake} along with the dataset.

\begin{table*}[h]
    \centering
    \small
    \renewcommand{\arraystretch}{1.1}
    \begin{tabular}{c|cc|c|c}
        \textbf{Task} & \# of \textbf{correct$_{ans}$} traces & \# of \textbf{incorrect$_{ans}$} traces & \# of \textbf{incorrect$_{mis}$} traces & \textbf{Total}\\
        \hline
        Word sorting & 45 & 255 & 266 & 300 \\
        Tracking shuffled objects & 45 & 255 & 260 & 300 \\
        Logical deduction & 45 & 255 & 294 & 300 \\
        Multistep arithmetic & 45 & 255 & 238 & 300 \\
        Dyck languages & 482 & 504 & 650 & 986 \\
        Dyck languages (sampled) & 88 & 504 & 545 & 592 \\
    \end{tabular}
    \vspace{-0.2cm}
    \caption{Number of traces in our dataset that are correct and incorrect. Dyck languages (sampled) is a set of traces sampled so that the ratio of correct$_{ans}$ to incorrect$_{ans}$ traces matches other tasks.}
    \label{tab:dataset}
    \vspace{-0.5cm}
\end{table*}

\subsection{Annotation}
Each generated trace is annotated with the first logical error. We ignore any subsequent errors as they may be dependent on the original error.

Note that traces can contain a logical mistake yet arrive at the correct answer. To disambiguate the two types of correctness, we will use the terms \textbf{correct$_{ans}$} and \textbf{incorrect$_{ans}$} 
to refer to whether the final \textbf{ans}wer of the trace is correct. \textbf{Accuracy$_{ans}$} would therefore refer to the overall accuracy for the task, based on how many final answers are correct. To refer to whether the trace contains a logical \textbf{mis}take (rather than the correctness of the final answer), we will use \textbf{correct$_{mis}$} and \textbf{incorrect$_{mis}$}.

\subsubsection{Human annotation}
For 4 of the 5 tasks, we recruit human annotators to go through each trace and identify any errors. Annotators have no domain expertise but are given guidelines\footnote{\url{https://github.com/WHGTyen/BIG-Bench-Mistake} contains further details. \label{footnote:url}} to complete the task.

Before annotation, we sample a set of 300 traces for each task, where 255 (85\%) are incorrect$_{ans}$, and 45 (15\%) are correct$_{ans}$. Since human annotation is a limited and expensive resource, we chose this distribution to maximise the number of steps containing mistakes and to prevent over-saturation of correct steps. We also include some correct$_{ans}$ traces because some may contain logical errors despite the correct answer, and to ensure that the dataset included examples of correct steps that are near the end of the trace. To account for this skewed distribution, results in \autoref{sec:backtracking} are split according to whether the original trace is correct$_{ans}$ or not.

Following \citet{lightman2023let}, annotators are instructed to go through each step in the trace and indicate whether the step is correct or not (binary choice). Annotators only submit their answers when all steps are annotated, or there is one incorrect step. If an incorrect step is identified, the remaining steps are skipped. This is to avoid ambiguities where a logically correct deduction is dependent on a previous mistake. Our annotation guidelines can be found at \url{https://github.com/WHGTyen/BIG-Bench-Mistake/tree/main/annotation_guidelines}, and we include a screenshot of the user interface in \autoref{sec:ui}.

Each trace is annotated by at least 3 annotators. If there are any disagreements, we take the majority label. We calculate Krippendorff's alpha \citep{hayes2007answering} to measure inter-rater reliability (see \autoref{tab:krippendorff}). 

\begin{table}[h]
    \centering
    \small
    \begin{tabular}{rl}
        \textbf{Task} & \textbf{Krippendorff's $\alpha$}\\
        \hline
        Word sorting & 0.979 \\
        Tracking shuffled objects & 0.998 \\
        Logical deduction & 0.996 \\
        Multistep arithmetic & 0.984 \\
    \end{tabular}
    \caption{Inter-rater reliability for the human-annotated tasks, measured by Krippendorff's alpha.}
    \label{tab:krippendorff}
\end{table}

\subsubsection{Automatic annotation}
For Dyck languages, we use mostly automatic instead of human annotation, as the traces show limited variation in phrasing and solution paths.

For each trace, we algorithmically generate a set of steps based on the format used in the prompt examples. Using pattern matching, we identify whether each model-generated step conforms to the same format. If so, we compare the two and assume that the trace is incorrect if the symbols do not match. Additionally, we account for edge cases such as where the final two steps are merged into one, or variations in presentation where symbols may or may not be placed in quotes. We release the code at \url{https://github.com/WHGTyen/BIG-Bench-Mistake} along with our dataset.


\section{Can LLMs \textit{find} reasoning mistakes in CoT traces?}
\label{sec:benchmark_results}
\autoref{tab:benchmark_results} shows the accuracy of GPT-4-Turbo, GPT-4, GPT-3.5-Turbo, Gemini Pro, and PaLM 2 Unicorn on our mistake-finding dataset. For each question, the possible answers are either: that there are no mistakes, or; if there is a mistake, the number N indicating the step in which the first mistake occurs. A model's output is only considered correct if the location matches exactly, or the output correctly indicates that there are no mistakes.

All models are given the same 3-shot prompts\cref{footnote:url}. We use three different prompting methods:

\begin{itemize}[topsep=0.2cm,itemsep=0cm,leftmargin=0.5cm]
    \item \textbf{Direct trace-level prompting} involves using the whole trace as input to the model and directly prompting for the mistake location. The model must output either the number representing the step, or "No".
    
    \item \textbf{Direct step-level prompting} prompts for a binary Yes/No output for every step, indicating whether or not the step is correct. In each generation call, the input contains the partial trace up to (and including) the target step, but does not contain results for previous steps. The final answer is inferred from where the first "No" output occurs (subsequent steps are ignored).
    
    \item \textbf{CoT step-level prompting} is an extension of direct, step-level prompting. Instead of a binary Yes/No response, we prompt the model to check the (partial) trace through a series of reasoning steps. This method is the most resource intensive of all three methods as it involves generating a whole CoT sequence for every step. As with direct step-level prompting, the final answer is inferred from where the first "No" output occurs (subsequent steps are ignored).
\end{itemize}

\subsection{Discussion}
All five models appear to struggle with our mistake finding dataset. GPT-4 attains the best results but only reaches an overall accuracy of 52.87 with direct step-level prompting. While exact parameter counts are undisclosed, GPT-4 is likely one of the largest models, along with PaLM 2 Unicorn\footnote{Note that the traces in our dataset are generated using PaLM 2 Unicorn and are sampled according to whether the final answer was correct or not. Therefore, we expect that using PaLM 2 itself to do mistake finding will produce different and likely biased results. Further work is needed to elucidate the difference between cross-model evaluation and self-evaluation.}, while Gemini Pro and GPT-3.5-Turbo are among the smaller models.

Our findings are in line with and builds upon results from \citet{huang2023large}, who show that existing self-correction strategies are ineffective on reasoning errors. In our experiments, we specifically target the models' \textit{mistake finding} ability and provide results for additional tasks. We show that state-of-the-art LLMs clearly struggle with mistake finding, even in the most simple and unambiguous cases. (For comparison, humans can identify mistakes without specific expertise, and have a high degree of agreement, as shown in \autoref{tab:krippendorff}.)

We hypothesise that LLMs' inability to find mistakes is a main contributing factor to why LLMs are unable to self-correct reasoning errors. If LLMs are unable to \textit{identify} mistakes, it should be no surprise that they are unable to self-correct either.

\begin{table}[htbp!]
    \centering
    \setlength\tabcolsep{-1pt}
    \begin{tabular}{r>{\centering\arraybackslash}m{1.7cm}>{\centering\arraybackslash}m{1.7cm}>{\centering\arraybackslash}m{1.6cm}}
    \hline
     \textbf{Model} & \textbf{Direct} (trace) & \textbf{Direct} (step) & \textbf{CoT} (step) \\
    \hline
    \multicolumn{4}{c}{\textbf{Word sorting} (11.7)} \\
     GPT-4-Turbo    & 36.33 & 33.00 & --    \\
     GPT-4          & 35.00 & 44.33 & 34.00 \\
     GPT-3.5-Turbo  & 11.33 & 15.00 & 15.67 \\
     Gemini Pro     & 10.67 & --    & --    \\
     PaLM 2 Unicorn      & 11.67             & 16.33             & 14.00            \\
    \hline
    \multicolumn{4}{c}{\textbf{Tracking shuffled objects} (5.4)} \\
     GPT-4-Turbo    & 39.33 & 61.67 & --    \\
     GPT-4          & 62.29 & 65.33 & 90.67 \\
     GPT-3.5-Turbo  & 10.10 & 1.67  & 19.00 \\
     Gemini Pro     & 37.67 & --    & --    \\
     PaLM 2 Unicorn      & 18.00             & 28.00             & 55.67            \\
    \hline
    \multicolumn{4}{c}{\textbf{Logical deduction} (8.3)} \\
     GPT-4-Turbo & 21.33 & 75.00 & -- \\
     GPT-4          & 40.67             & 67.67              & 10.33             \\
     GPT-3.5-Turbo   & 2.00             & 25.33             & 9.67            \\
     Gemini Pro     & 8.67 & --    & --    \\
     PaLM 2 Unicorn      & 6.67             & 38.00             & 12.00            \\
    \hline
    \multicolumn{4}{c}{\textbf{Multistep arithmetic} (5.0)} \\
     GPT-4-Turbo    & 38.33 & 43.33 & -- \\
     GPT-4          & 44.00             & 42.67              & 41.00             \\
     GPT-3.5-Turbo   & 20.00             & 26.00             & 25.33            \\
     Gemini Pro     & 21.67 & --    & --    \\
     PaLM 2 Unicorn      & 22.00             & 21.67             & 23.67            \\
    \hline
    \multicolumn{4}{c}{\textbf{Dyck languages\textdagger} (24.5)} \\
     GPT-4-Turbo    & 15.33* & 28.67* & -- \\
     GPT-4          & 17.06             & 44.33*              & 41.00*             \\
     GPT-3.5-Turbo   & 8.78             & 5.91             & 1.86            \\
     Gemini Pro     & 2.00 & --    & --    \\
     PaLM 2 Unicorn      & 10.98             & 14.36             & 17.91           \\
    \hline
    \multicolumn{4}{c}{\textbf{Overall}} \\
     GPT-4-Turbo    & 30.13 & 48.33 & -- \\
     GPT-4          & 39.80             & 52.87              & 43.40             \\
     GPT-3.5-Turbo   & 10.44             & 14.78             & 14.31            \\
     Gemini Pro     & 16.14 & --    & --    \\
     PaLM 2 Unicorn    & 17.09 & 23.67 & 24.65 \\
    \hline
    \end{tabular}
    \vspace{-0.2cm}
    \caption{Mistake finding accuracy across 5 tasks. The average number of steps 
    in CoT reasoning traces in each task is in brackets. Unless otherwise indicated, the number of traces is in \autoref{tab:dataset}. We provide scores split by correctness$_{ans}$ of the original trace in \autoref{sec:split_scores}. Due to cost and usage limits, we are unable to provide results indicated by --.
    \\\hspace{\textwidth}\textdagger~indicates that traces were sampled to contain 15\% correct$_{ans}$ and 85\% incorrect$_{ans}$ traces (see \autoref{tab:dataset}).
    \\\hspace{\textwidth}* indicates that traces were sampled to contain 45 correct$_{ans}$ and 255 incorrect$_{ans}$ traces to reduce costs.}
    \label{tab:benchmark_results}
    \vspace{-0.6cm}
\end{table}

\subsection{Comparison of prompting methods}
\begin{figure}[!ht]
    \centering
    \includegraphics[width=0.45\textwidth,trim={0 0cm 1cm 1cm},clip]{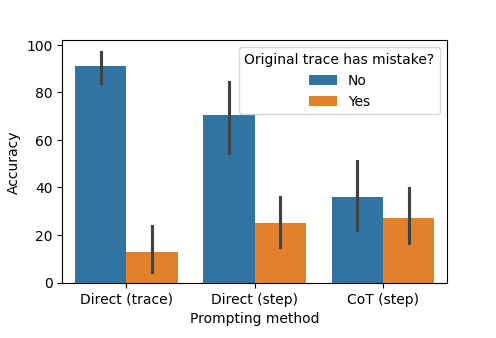}
    \vspace{-0.3cm}
    \caption{Graph of mistake location accuracies 
    for each prompting method (excluding GPT-4-Turbo and Gemini Pro which we do not have all results for). Blue bars show accuracies on traces with no mistakes, so the model must predict that the trace has no mistake to be considered correct; orange bars show accuracies on traces with a mistake, so the model must predict the precise location of the mistake to be considered correct.}
    \label{fig:tradeoff}
\end{figure}

As we compare results across the three methods, we find that the accuracy on traces with no mistakes goes down\footnote{Note that the traces in BIG-Bench Mistake are sampled to contain more incorrect$_{ans}$ traces than correct$_{ans}$ traces (and therefore more incorrect$_{mis}$ traces than correct$_{mis}$ traces), so the overall mistake location accuracy appears higher for per-step prompting in \autoref{tab:benchmark_results}, despite the poor accuracy for correct$_{mis}$ traces. For a full set of scores split by correctness$_{mis}$, see \autoref{sec:split_scores}.} considerably from direct, trace-level prompting to CoT, step-level prompting. \autoref{fig:tradeoff} demonstrates this trade-off.

We hypothesise that this is due to the number of outputs generated by the model. Our three methods involve generating increasingly complex outputs, starting with direct, trace-level prompting requiring a single token, then direct, step-level prompting requiring one token per step, and finally CoT step-level prompting requiring several sentences per step. If each generation call has some probability of identifying a mistake, then the more calls made on each trace, the more likely the model will identify at least one mistake.

\subsection{Few-shot prompting for mistake location as a proxy for correctness}
In this section, we investigate whether our prompting methods can reliably determine the correctness$_{ans}$ of a trace rather than the mistake location. Our motivation was that even humans use mistake finding as a strategy for determining whether an answer is correct or not, like when going through mathematical proofs or argumentation. Additionally, it may be the case that directly predicting the correctness$_{ans}$ of a trace is easier than pinpointing the precise location of an error.

\begin{table}[h!]
    \centering
    \setlength\tabcolsep{-1pt}
    \begin{tabular}{r>{\centering\arraybackslash}m{1.7cm}>{\centering\arraybackslash}m{1.7cm}>{\centering\arraybackslash}m{1.6cm}}
    \hline
     \textbf{Model} & \textbf{Direct} (trace) & \textbf{Direct} (step) & \textbf{CoT} (step) \\
    \hline
    \multicolumn{4}{c}{\textbf{Word sorting}} \\
     GPT-4-Turbo & 87.73 & 86.68 & -- \\
     GPT-4          & 81.50             & 85.12              & 81.19             \\
     GPT-3.5-Turbo   & 6.58             & 35.07             & 77.79            \\
     Gemini Pro      & 69.93 & -- & -- \\
     PaLM 2 Unicorn      & 21.08             & 56.66             & 62.92            \\
    \hline
    \multicolumn{4}{c}{\textbf{Tracking shuffled objects}} \\
     GPT-4-Turbo    & 52.23 & 74.31 & -- \\
     GPT-4          & 76.38             & 75.69              & 95.03             \\
     GPT-3.5-Turbo   & 32.04             & 77.61             & 78.11            \\
     Gemini Pro      & 79.66 & -- & -- \\
     PaLM 2 Unicorn      & 22.18             & 48.77             & 78.29            \\
    \hline
    \multicolumn{4}{c}{\textbf{Logical deduction}} \\
     GPT-4-Turbo & 86.46 & 81.79 & -- \\
     GPT-4          & 84.54             & 83.38              & 23.96             \\
     GPT-3.5-Turbo      & 10.34             & 67.62             & 61.31            \\
     Gemini Pro      & 48.57 & -- & -- \\
     PaLM 2 Unicorn   & 31.67             & 37.93             & 21.21            \\
    \hline
    \multicolumn{4}{c}{\textbf{Multistep arithmetic}} \\
     GPT-4-Turbo    & 71.17 & 86.24 & -- \\
     GPT-4          & 72.97             & 78.67              & 79.67             \\
     GPT-3.5-Turbo   & 3.76             & 53.18             & 64.08             \\
     Gemini Pro      & 32.21 & -- & -- \\
     PaLM 2 Unicorn      & 33.69             & 13.42             & 70.94            \\
    \hline
    \multicolumn{4}{c}{\textbf{Dyck languages}} \\
     GPT-4-Turbo    & 51.96 & 85.87 & -- \\
     GPT-4          & 62.33             & 85.73              & 79.60             \\
     GPT-3.5-Turbo   & 46.57             & 79.31             & 77.79            \\
     Gemini Pro      & 61.24 & -- & -- \\
     PaLM 2 Unicorn      & 31.17             & 31.63             & 25.20           \\
    \hline
    \end{tabular}
    \vspace{-0.2cm}
    \caption{Weighted average F1 scores for predicted correctness$_{ans}$ of traces across 5 tasks. Baseline is 78 if we only select the incorrect$_{ans}$ label. As in \autoref{tab:benchmark_results}, traces for the Dyck languages task has been sampled to match the ratio of correct$_{ans}$ to incorrect$_{ans}$ traces of the other tasks. See \autoref{tab:dataset} for a full breakdown. 
    }
    \label{tab:correctness_proxy}
    \vspace{-0.5cm}
\end{table}

We calculate averaged F1 scores based on whether the model predicts there is a mistake in the trace. If there is a mistake, we assume the model prediction is that the trace is incorrect$_{ans}$. Otherwise, we assume the model prediction is that the trace is correct$_{ans}$. In \autoref{tab:correctness_proxy}, we average the F1s calculated with correct$_{ans}$ and incorrect$_{ans}$ as positive labels, weighted according to the number of times each label occurs. Note that the naive baseline of predicting all traces as incorrect achieves a weighted F1 average of 78.

The weighted F1 scores show that prompting for mistakes is likely a poor strategy for determining the correctness of the final answer. This is in line with our previous finding that LLMs struggle to identify mistake locations, and also builds upon results from \citet{huang2023large}, who demonstrate that improvements from Reflexion \citep{shinn2023reflexion} and RCI \citep{kim2023language} are only from using oracle correctness$_{ans}$ information.

\section{Can LLMs \textit{correct} reasoning mistakes in CoT traces?}
\label{sec:backtracking}
In this section, we examine LLMs' ability to \textit{correct} mistakes, independently of their ability to \textit{find} them. To do so, we feed oracle mistake location information from BIG-Bench Mistake into the model and prompt it for a corrected version of the original CoT trace.

\begin{figure*}[!htb]
    \centering
    \includegraphics[width=0.75\textwidth]{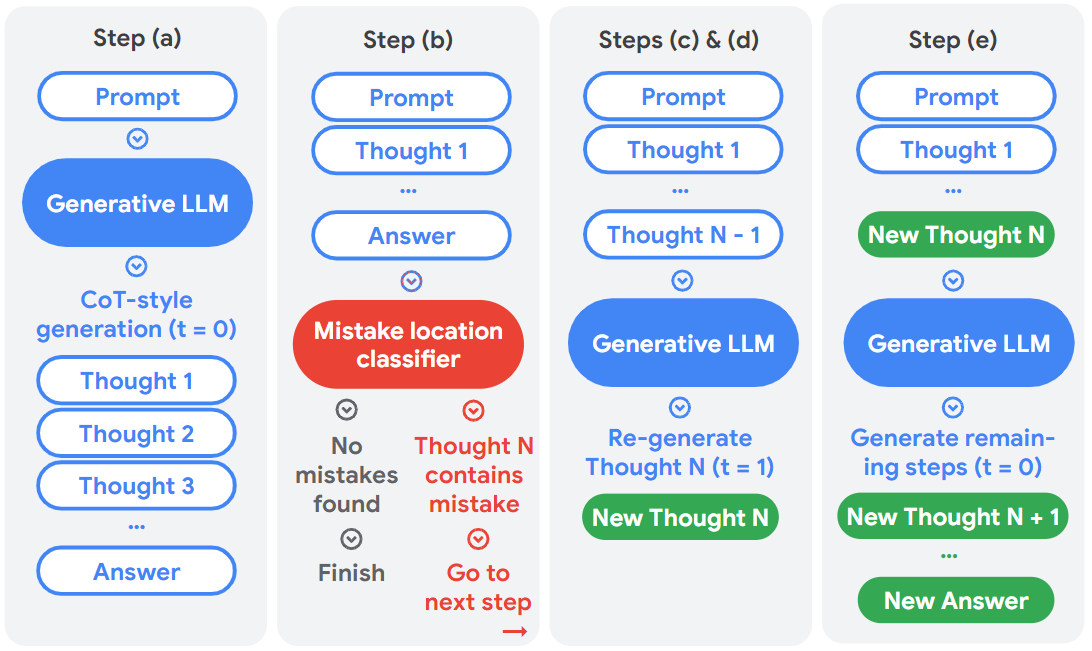}
    \caption{Visualization of our backtracking method, which is used to feed mistake location information to the model for correction. $t$ refers to the temperature used during generation.}
    \label{fig:backtracking}
\end{figure*}

As a simple baseline, we use the following backtracking method (visualized in \autoref{fig:backtracking}):
\begin{enumerate}[topsep=0.1cm,itemsep=-0.05cm,leftmargin=0.5cm,label=(\alph*)]
    \item First, the model generates an initial CoT trace. In our experiments, we use temperature $= 0$.
    \item We then determine the mistake location in this trace, either from oracle labels (in this section) or with a classifier (in \autoref{sec:classifier}).
    \item If there are no mistakes, we move onto the next trace. If there is a mistake (e.g. at Thought 4 in the example trace in \autoref{tab:example}), we prompt the model again for the same step but at temperature $= 1$. We use same prompt and the partial trace containing all steps up to but not including the mistake step (e.g. up to Thought 3, prompting for Thought 4).
    \item In our experiments, we found that (c) often produced steps that are identical to the original. We therefore repeat (c) until a different step is generated (or up to a fixed number, whichever is less). For this paper, we use 8 as the maximum number of re-generations; the effects of varying this number is left for future investigation. To reduce computational cost, we generate 8 outputs simultaneously but only select one for backtracking.
    \item Finally, with the new, regenerated step in place of the previous one, we generate the remaining steps of the trace again at temperature $= 0$.
\end{enumerate}

This backtracking method is designed to be a very simple baseline, with no specific prompt text or phrasing, and without relying on generating a large number of alternatives. For our experimental results below, we specifically use the same model (PaLM 2 Unicorn) to correct the traces it originally generated, to test its ability to self-correct.

\begin{table*}[!htbp]
    \centering
    \begin{tabular}{rcccc>{\centering\arraybackslash}m{1.8cm}}
        \hline
         & \multicolumn{2}{c}{With \textbf{mistake} location} & \multicolumn{2}{c}{With \textbf{random} location} & \multirow{2}{1.8cm}{Avg. num. of steps} \\
        \cline{2-5}
         \textbf{Task} & \textbf{$\Delta$ accuracy $_{\cmark}$} & \textbf{$\Delta$accuracy$_{\xmark}$} & \textbf{$\Delta$ accuracy $_{\cmark}$} & \textbf{$\Delta$accuracy$_{\xmark}$} & \\
        \hline
         Word sorting              & -11.11       & +23.53         & -15.56       & +11.76         & 11.7\\
         Tracking shuffled objects & -6.67        & +43.92         & -6.67       & +20.39         & 5.4\\
         Logical deduction         & -11.43       & +36.86         & -13.33                   & +21.57                     & 8.3\\
         Multistep arithmetic      & -0.00       & +18.04         & -8.89                   & +10.59                     & 5.0\\
         Dyck languages            & -6.82        & +18.06         & -15.91                   & +5.16                     & 24.5\\
        \hline
    \end{tabular}
    \caption{Absolute differences in accuracy$_{ans}$ before and after backtracking. "With mistake location" indicates that backtracking was done using oracle mistake locations from the dataset; "With random location" indicates that backtracking was done based on randomly selected locations. \textbf{$\Delta$accuracy$_{\cmark}$} refers to differences in accuracy$_{ans}$ on the set of traces whose \textit{original} answer was correct$_{ans}$; \textbf{$\Delta$accuracy$_{\xmark}$} for traces whose original answer was incorrect$_{ans}$. The average number of steps in a trace is shown to demonstrate the likelihood of randomly selecting the correct mistake location in the random baseline condition.}
    \label{tab:backtracking}
    \vspace*{-0.3cm}
\end{table*}

\subsection{Results}
The results are shown in \autoref{tab:backtracking}. To show that performance increases are not due to randomly resampling outputs, we compare our results to a random baseline, where a mistake location\footnote{As described above, the mistake location can be either the number representing the step, or that there are no mistakes. If there are no mistakes, we do not use backtracking and simply use the original trace.} is randomly selected for each trace and we perform backtracking based on the random location.

Note that \autoref{tab:backtracking} separates results into numbers for the correct set and the incorrect set, referring to whether the \textit{original} trace was correct$_{ans}$ or not. This gives a clearer picture than the overall accuracy$_{ans}$, which would be skewed by the proportion of traces that were originally correct$_{ans}$ (15\%) and incorrect$_{ans}$ (85\%).

Scores represent the absolute differences in accuracy$_{ans}$. We perform backtracking on both correct$_{ans}$ and incorrect$_{ans}$ traces, as long as there is a mistake in one of the steps.

\paragraph{$\Delta$accuracy$_{\cmark}$} refers to differences in accuracy$_{ans}$ on the set of traces whose \textit{original} answer was correct$_{ans}$. Note that we take losses here because, despite the correct answer, there is a logical mistake in one of the steps. Therefore, the answer may change to an incorrect one when we backtrack.
\vspace{-0.2cm}
\paragraph{$\Delta$accuracy$_{\xmark}$} is the same but for incorrect$_{ans}$ traces, so the answers may have been corrected, hence increasing accuracy$_{ans}$.

For example, for the word sorting task, 11.11\% of traces that were originally correct$_{ans}$ became incorrect$_{ans}$, while 23.53\% of traces that were originally incorrect$_{ans}$ became correct$_{ans}$.

\subsection{Discussion}
The scores show that the gains from correcting incorrect$_{ans}$ traces are larger than losses from changing originally correct answers. Additionally, while the random baseline also obtained improvements, they are considerably smaller than if the true mistake location was used. Note that tasks involving fewer steps are more likely to improve performance in the random baseline, as the true mistake location is more likely to be identified.

Our results show that, with mistake location information available, LLMs can correct their own outputs and improve overall downstream performance. This suggests that the main bottleneck in self-correction methods is the identification of mistakes, rather than the correction process. This bottleneck can be overcome by using ground truth feedback (as in Reflexion \citep{shinn2023reflexion} or RCI \citep{kim2023language}), or by training a classifier (see \autoref{sec:classifier}).

While our numbers do show that our gains are higher than our losses, it should be noted that changes in the overall accuracy depends on the original accuracy achieved on the task. For example, if the original accuracy on the tracking shuffled objects task was 50\%, the new accuracy would be 68.6\%. On the other hand, if the accuracy was 99\%, the new accuracy would drop to 92.8\%. As our dataset is highly skewed and only contains 45 correct$_{ans}$ traces per task, we leave to future work a more comprehensive assessment of backtracking, as well as the development of more sophisticated ways to incorporate mistake location information into the self-correction loop.

\section{Obtaining mistake location information with a trained classifier}
\label{sec:classifier}
As shown in \autoref{sec:backtracking}, if mistake location information is available, LLMs can correct their own CoT traces and boost downstream performance. However, these experimental results are based on oracle labels, which are typically not available in downstream tasks.

One possible solution is to obtain mistake location information from a smaller, trained classifier. If training data is available, one might ask why this approach is preferable to simply fine-tuning the larger, generator model. The reasons are:
\begin{itemize}[topsep=0.1cm,itemsep=-1pt,leftmargin=0.5cm]
    \item Training a small classifier is far more efficient in terms of computing resources and available data.
    \item Once the classifier is trained, it can be used with any LLM as the generator and be updated independently. This can be especially helpful with API-based LLMs that cannot be fine-tuned.
    \item The process of mistake finding is more interpretable than updating the weights of the generator model directly. It clearly pinpoints the location at which an error occurs, which can help the debugging process and allow faster development and iterations of models.
\end{itemize}

\vspace{0.2cm}
In this section, we seek to answer two questions in the following subsections:

\paragraph{\textbf{5.1: What mistake-finding accuracy is required for backtracking to be effective?}} \mbox{} \\
A trained classifier is unlikely to reach 100\% mistake-finding accuracy. If backtracking is only effective when mistake location is 100\% accurate, we would not be able to replace oracle labels with a trained classifier.
\paragraph{\textbf{5.2: Is it possible to improve on results in \autoref{sec:benchmark_results} without in-domain training data?}} \mbox{} \\
Sufficient in-domain training data typically guarantees a performance boost, but can be hard to obtain. We investigate whether mistake-finding in reasoning traces is transferable across tasks. If so, one can use BIG-Bench Mistake or similar datasets to fine-tune a mistake-finding classifier for other tasks.

\subsection{Minimum mistake finding accuracy} \label{sec:simulated_classifier}
To explore what level of mistake-finding accuracy is needed, we simulate classifiers at different levels of accuracy and run backtracking for each level. We use accuracy$_{mis}$ to refer to the mistake-finding accuracy classifier, to differentiate from downstream task accuracy$_{ans}$.

For a given classifier at $X$\% accuracy$_{clf}$, we use the mistake location from BIG-Bench Mistake $X$\% of the time. For the remaining $(100 - X)$\%, we sample a mistake location randomly. To mimic the behaviour of a typical classifier, mistake locations are sampled to match the distribution found in the dataset. We also ensure that the sampled location does not match the correct location.

\begin{figure}[h]
    \centering
    \includegraphics[width=0.5\textwidth,trim={0 0.3cm 0 0.5cm},clip]{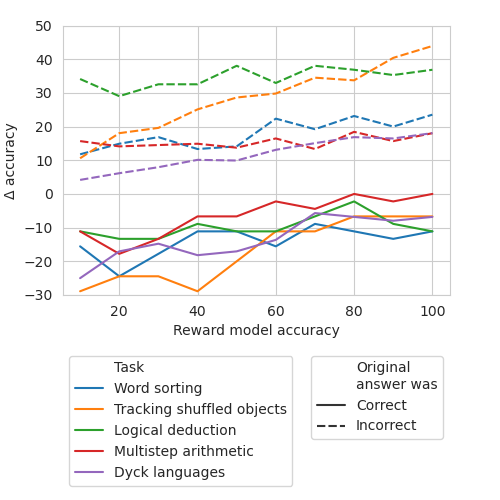}
    \caption{$\Delta$accuracy$_{\cmark}$ and $\Delta$accuracy$_{\xmark}$ on each dataset as accuracy$_{clf}$ increases.}
    \label{fig:simulated_rm}
    \vspace{-0.2cm}
\end{figure}

Results are shown in \autoref{fig:simulated_rm}. We can see that the losses in $\Delta$accuracy$_{\cmark}$ begins to plateau at 65\%. In fact, for most tasks, $\Delta$accuracy$_{\cmark}$ is already larger than $\Delta$accuracy$_{\xmark}$ at around 60-70\% accuracy$_{RM}$. This demonstrates that while higher accuracies produce better results, backtracking is still effective even without gold standard mistake location labels.

\subsection{Training a classifier on out-of-domain data} \label{sec:reward_modeling}
We test whether mistake-finding can benefit from a dedicated classifier trained on out-of-distribution tasks. We fine-tune PaLM 2 Otter, a model much smaller than PaLM 2 Unicorn, with our BIG-Bench Mistake data for 20k steps and choose the checkpoint with the best validation results. For each task, we hold out the in-domain data for evaluation while training the classifier on the other 4 tasks. 
\begin{table*}[h]
    \centering
    \small
    \begin{tabular}{c|>{\centering\arraybackslash}m{2.5cm}>{\centering\arraybackslash}m{3cm}|c}
         Held-out task & Trained classifier accuracy$_{mis}$ (Otter) & 3-shot prompting accuracy$_{mis}$ (Unicorn) & Difference \\
         \hline
         Word sorting              & \textbf{22.33} & 11.67 & +11.66  \\
         Tracking shuffled objects & \textbf{37.67} & 18.00 & +19.67  \\
         Logical deduction         & 6.00  & \textbf{6.67}  & -0.67   \\
         Multi-step arithmetic     & \textbf{26.00} & 22.00 & +4.00    \\
         Dyck languages            & \textbf{33.57} & 10.98 & +22.59  \\
    \end{tabular}
    \caption{Absolute difference in mistake finding accuracy between PaLM 2 Unicorn and a small, trained classifier. Bold indicates the best score for each task. Note that PaLM 2 Otter is significantly smaller than PaLM 2 Unicorn, and is trained on out-of-domain data.}
    \label{tab:rewardmodelexternal}
    \vspace*{-0.5cm}
\end{table*}

We show the relative improvements and losses in~\autoref{tab:rewardmodelexternal} vs. a 3-shot baseline on PaLM 2 Unicorn (scores from \autoref{sec:benchmark_results}). We see gains for 4 out of 5 of the tasks. Note the classifiers we train are significantly smaller than our inference model, and is trained on out-of-domain data. This suggests that it may be possible to train classifiers to assist in backtracking, and that these classifiers do not have to be large. Further, such a classifier can work on out-of-distribution mistakes.

Despite this, the performance of our classifiers do not meet the threshold required for effective backtracking, as demonstrated in \autoref{sec:simulated_classifier}. We believe more data may be necessary to improve results across the board on all tasks. We leave to future work the collection of this larger dataset and a more rigorous investigation of the trade-offs of model size vs. performance of the classifier.

We also leave for future investigation the effect of backtracking iteratively with a classifier: for example, the generator model may make another mistake after backtracking for the first time, which can then be identified and corrected again.

\section{Related work}

\subsection{Datasets}
To our knowledge, the only publicly available dataset containing mistake annotations in LLM outputs is PRM800K \citep{lightman2023let}, which is a dataset of solutions to Olympiad-level math questions. Our dataset BIG-Bench Mistake covers a wider range of tasks to explore the reasoning capabilities of LLMs more thoroughly. Additionally, our dataset explores the more general case of prompting API-based LLMs, whereas PRM800K uses a generator LLM that was heavily fine-tuned for math-specific problems.

\subsection{Self-correction}
The term \textit{self-correction} can be applied to a wide variety of techniques. \citet{pan2023automatically} present a survey of self-correction methods in recent literature. Broadly, proposed self-correction methods can vary in the following dimensions:

\paragraph{Source of feedback} Some techniques rely on external feedback inherent to the task such as code execution errors \citep{yasunaga2020graph,chen2023codet,chen2024teaching}, or feedback from humans \citep[e.g.][]{chen2024learning,yuan-etal-2024-system}. Some explicitly train a model to produce feedback \citep[e.g.][]{madaan-etal-2021-think,welleck2023generating,bai2022constitutional,lee2023rlaif,paul2023refiner,ouyang2022training,bai2022training,ganguli2023capacity}, while others rely on prompting only \citep[e.g.][]{shinn2023reflexion,miao2024selfcheck}. It has been demonstrated that prompting-only setups can work well for stylistic or qualitative improvements \citep[e.g.][]{madaan2023self,chen2023iterative}, but would require external feedback for reasoning tasks \citep{huang2023large,shinn2023reflexion,kim2023language,madaan2023self}.

\paragraph{Time of correction} Feedback can be incorporated at various points during the self-correction process. Some methods do so by updating weights during training time \citep[e.g.][]{ouyang2022training,bai2022training,ganguli2023capacity}; some do so during generation time \citep[e.g.][]{weng-etal-2023-large,dalvi-mishra-etal-2022-towards,xie2023selfevaluation}; others apply correction to output that has already been generated \citep[e.g.][]{saunders2022self,kim2023language,shinn2023reflexion}. Our method falls into the final category (post-hoc correction), as it involves identifying an incorrect step in a complete CoT trace; however, it is also possible to apply a mistake-finding classifier at every step during generation.

\section{Conclusion}
In this paper, we investigate LLMs' ability to find mistakes and correct outputs. We find that LLMs generally struggle to find mistakes, but, when given mistake location information, are able to correct outputs to boost performance. We therefore hypothesise that mistake finding is an important bottleneck preventing self-corrections strategies from performing well on reasoning tasks.

We show initial evidence that a dedicated classifier for mistake finding can overcome this bottleneck. We train a small, baseline classifier on out-of-domain data and demonstrate improvement over few-shot prompting results. While the classifier does not reach the threshold required for effective backtracking, our results show that it is possible to improve on mistake-finding accuracy using standard machine learning techniques. We leave the development of more sophisticated methods to future work, and release our dataset BIG-Bench Mistake to encourage this direction of research.

\section*{Limitations}
One main limitation of our dataset is that it features tasks that are artificial and unrealistic for real-world applications. We made this choice to minimise ambiguity and subjectivity during the mistake finding process, but further work needs to be done to determine the effectiveness of backtracking in a more realistic setting.

Another limitation is that our paper does not experiment with backtracking on the original datasets on BIG-Bench, only showing results on the limited set that we sampled in a skewed manner, in order to maximise the value of the human annotators' time. We leave the full evaluation to future work as this is beyond the scope of this paper, which is intended as a proof-of-concept to show the importance of mistake-finding.


\section*{Acknowledgements}
We thank Vicky Zayats and Sian Gooding for their invaluable feedback, as well as members of the Google Research and Google DeepMind teams who gave continued support throughout the project. 

\bibliography{anthology,custom}
\bibliographystyle{acl_natbib}

\appendix
\section{Dataset details}
Our dataset, BIG-Bench Mistake, is available at \url{https://github.com/WHGTyen/BIG-Bench-Mistake} under the Apache License 2.0. The five tasks used in our dataset are based on BIG-Bench \citep{srivastava2022beyond}, also released under the Apache License 2.0. All five tasks are in the English language.

\subsection{3-shot CoT prompting to generate traces for BIG-Bench Mistake}
We use PaLM 2 (Unicorn) to generate the traces used in BIG-Bench Mistake. All traces are generated at temperature $= 0$.

Our prompts and examples can be found at \url{https://github.com/WHGTyen/BIG-Bench-Mistake}. Our prompts are based on chain-of-thought prompts in the BIG-Bench Hard dataset \citep{bigbenchhardprompts}, with four main changes:
\begin{enumerate}
    \item Example CoT traces in the prompt are broken up into smaller steps (typically one sentence per step). This is done so that mistake location information is more precise.
    \item Following \citet{yao2022react}, each step in the prompt is signposted with ``Thought 1'', ``Thought 2:'', etc. This allows us to refer to the number of the step when prompting for mistake location.
    \item For the logical deduction task, we find that the notation used in the original prompt with question marks is often inconsistent. It becomes difficult for annotators to determine whether a question mark is a mistake or not, because the correctness of the question mark is dependent on its interpretation. To minimise such ambiguity, we replace the question mark notation with text descriptions of the objects.
    \item For the multistep arithmetic task, one of the prompt examples is altered to increase the length of the equation. This is because the BIG-Bench Hard dataset (where the prompts are taken from) only used equations of a specific length, but our dataset contains equations of averaged a variety of lengths, in accordance with the original BIG-Bench dataset \citep{srivastava2022beyond}.
\end{enumerate}

Following \citet{yao2022react}, we use the newline as the stop token, thereby generating one step with every generation call. We algorithmically append ``Thought N:'' before each step. This allows us to split up steps in a clear and systematic way. We stop generating once an answer is reached, which is detected using the following regex:\\
\verb|(?<=[Tt]he answer is).*$|

\subsection{3-shot prompting to identify mistakes in BIG-Bench Mistake}
As described in \autoref{sec:benchmark_results}, we explore three different methods of prompting for mistake location: direct trace-level prompting, direct step-level prompting, and CoT step-level prompting. We use 3-shot prompting for all methods, and our prompts and examples can be found at \url{https://github.com/WHGTyen/BIG-Bench-Mistake}.

Our prompts follow OpenAI's chat completion format. All results were obtained with temperature $= 0$ and no stop tokens. 

\section{Annotation}
We release our annotation guidelines at \url{https://github.com/WHGTyen/BIG-Bench-Mistake}. Our annotators are recruited via our institution and contracted at the market rate in their country of residence.

During annotation of the multistep arithmetic task, we found that the first CoT step given in the original BIG-Bench Hard prompt examples \citep{bigbenchhardprompts} was incorrect. Since all generated traces contained the same first step, we removed that step before showing traces to the annotators.

\begin{figure*}[!hb]
    \centering
    \includegraphics[width=\textwidth]{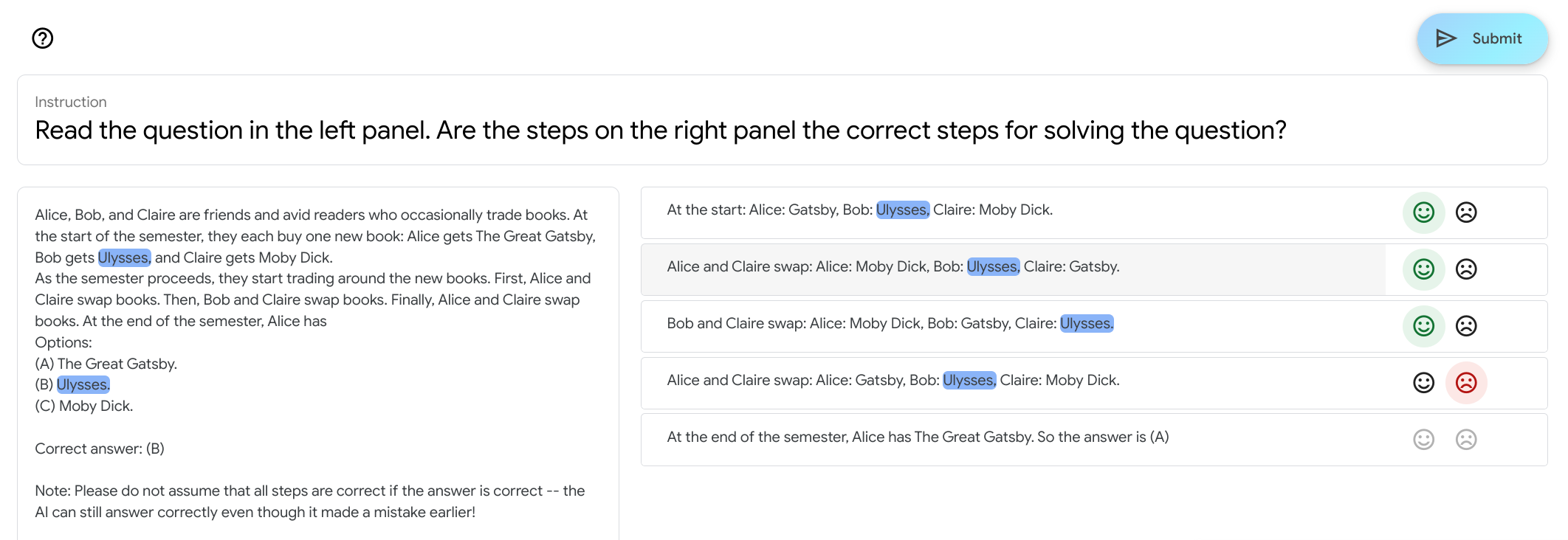}
    \caption{Screenshot of the user interface for a question from the tracking shuffled objects task.}
    \label{fig:ui}
\end{figure*}

\autoref{fig:ui} contains an example screenshot of the user interface. For every trace, we provide the input question as well as the target answer, with a note to be aware of errors that may occur in correct$_{ans}$ traces.

Annotators can click on words to highlight the same word across the trace and the question text, which we found was particularly helpful for some tasks such as word sorting and tracking shuffled objects. Buttons on the right automatically become inactive if a previous step has been labelled as negative.

\section{Training mistake-finding classifiers}
To train our mistake-finding classifiers (see \autoref{sec:reward_modeling}), we fine-tune PaLM 2 Otter on 4 of our 5 tasks, holding out one task for evaluation. This is done for each of our 5 tasks.

All 5 models are fine-tuned for 20k steps with a batch size of 32. The learning rate is $1e^{-5}$ with a linear ramp and cosine decay. After 20k steps, we select the checkpoint with the best validation results. The number of steps trained for each model are shown in \autoref{tab:rewardmodel_steps}.

\begin{table}[h]
    \centering
    \begin{tabular}{c c}
         \textbf{Held-out task} & \textbf{Training steps} \\
         \hline
         Word sorting              & 6800  \\
         Tracking shuffled objects & 8000  \\
         Logical deduction         & 9000   \\
         Multi-step arithmetic      & 10000    \\
         Dyck languages            & 10000  \\
    \end{tabular}
    \caption{Number of training steps to fine-tune each classifier.}
    \label{tab:rewardmodel_steps}
\end{table}

All models are trained as a binary classifier on whether a CoT step is correct, given the task and previous steps. Due to the limited data, we include in training the CoT steps that occur after the first mistake step. These steps are considered incorrect for the purposes of training (despite not being human-annotated as such).

\onecolumn
\section{User interface}
\label{sec:ui}

\clearpage
\section{Benchmark scores}
\label{sec:split_scores}
\begin{table*}[!hb]
    \centering
    \setlength\tabcolsep{-1pt}
    \begin{subtable}[!htb]{0.45\textwidth}
        \begin{tabular}{r>{\centering\arraybackslash}m{1.7cm}>{\centering\arraybackslash}m{1.7cm}>{\centering\arraybackslash}m{1.7cm}}
            \hline
             \textbf{Model} & \textbf{Direct} (trace) & \textbf{Direct} (step) & \textbf{CoT} (step) \\
            \hline
            \multicolumn{4}{c}{\textbf{Word sorting}} \\
             GPT-4-Turbo    & 67.74  & 38.24 & --    \\
             GPT-4          & 88.24  & 82.35 & 58.82 \\
             GPT-3.5-Turbo  & 100.00 & 97.06 & 20.59 \\
             Gemini Pro     & 44.12  & --    & --    \\
             PaLM 2 Unicorn & 100.00 & 73.53 & 35.29 \\
            \hline
            \multicolumn{4}{c}{\textbf{Tracking shuffled objects}} \\
             GPT-4-Turbo    & 90.00  & 77.50 & --    \\
             GPT-4          & 82.50  & 82.50 & 80.00 \\
             GPT-3.5-Turbo  & 67.50  & 0.00  & 0.00  \\
             Gemini Pro     & 12.50  & --    & --    \\
             PaLM 2 Unicorn & 100.00 & 85.00 & 47.50 \\
            \hline
            \multicolumn{4}{c}{\textbf{Logical deduction}} \\
             GPT-4-Turbo    & 100.00 & 83.33  & --     \\
             GPT-4          & 100.00 & 100.00 & 0.00   \\
             GPT-3.5-Turbo  & 100.00 & 50.00  & 100.00 \\
             Gemini Pro     & 33.33  & --     & --    \\
             PaLM 2 Unicorn & 100.00 & 100.00 & 50.00 \\
            \hline
            \multicolumn{4}{c}{\textbf{Multistep arithmetic}} \\
             GPT-4-Turbo    & 57.69 & 40.32 & --     \\
             GPT-4          & 53.23 & 46.77 & 27.42  \\
             GPT-3.5-Turbo  & 96.77 & 79.03 & 58.06  \\
             Gemini Pro     & 83.87  & --    & --    \\
             PaLM 2 Unicorn & 83.87 & 93.55 & 29.03 \\
            \hline
            \multicolumn{4}{c}{\textbf{Dyck languages}} \\
             GPT-4-Turbo    & 96.42 & 30.00 & --     \\
             GPT-4          & 98.41 & 78.57 & 13.79  \\
             GPT-3.5-Turbo  & 95.74 & 4.76  & 0.00   \\
             Gemini Pro     & 0.00  & --    & --    \\
             PaLM 2 Unicorn & 100.00 & 80.95 & 19.05 \\
            \hline
        \end{tabular}
        \caption{Mistake finding accuracy for traces that do not contain mistakes (correct$_{mis}$).}
        \label{tab:split_scores_correct}
    \end{subtable}
    \hfill
    \begin{subtable}[!hbp]{0.45\textwidth}
        \begin{tabular}{r>{\centering\arraybackslash}m{1.6cm}>{\centering\arraybackslash}m{1.6cm}>{\centering\arraybackslash}m{1.7cm}}
            \hline
             \textbf{Model} & \textbf{Direct} (trace) & \textbf{Direct} (step) & \textbf{CoT} (step) \\
            \hline
            \multicolumn{4}{c}{\textbf{Word sorting}} \\
             GPT-4-Turbo    & 32.71  & 32.33 & --    \\
             GPT-4          & 28.20  & 39.47 & 30.83 \\
             GPT-3.5-Turbo  & 0.00   & 4.51  & 15.04 \\
             Gemini Pro     & 6.39   & --    & --    \\
             PaLM 2 Unicorn & 0.38   & 9.02  & 11.28 \\
            \hline
            \multicolumn{4}{c}{\textbf{Tracking shuffled objects}} \\
             GPT-4-Turbo    & 31.54  & 59.23 & --    \\
             GPT-4          & 59.14  & 62.69 & 92.31 \\
             GPT-3.5-Turbo  & 1.17   & 1.92  & 21.92 \\
             Gemini Pro     & 41.54  & --    & --    \\
             PaLM 2 Unicorn & 5.38   & 19.23 & 56.92 \\
            \hline
            \multicolumn{4}{c}{\textbf{Logical deduction}} \\
             GPT-4-Turbo    & 20.81  & 74.83  & --     \\
             GPT-4          & 39.46  & 67.01  & 10.54   \\
             GPT-3.5-Turbo  & 0.00   & 24.83  & 7.82 \\
             Gemini Pro     & 8.16  & --    & --    \\
             PaLM 2 Unicorn & 4.76  & 36.73 & 11.22 \\
            \hline
            \multicolumn{4}{c}{\textbf{Multistep arithmetic}} \\
             GPT-4-Turbo    & 34.27 & 44.12 & --     \\
             GPT-4          & 41.60 & 41.60 & 44.54  \\
             GPT-3.5-Turbo  & 0.00  & 12.18 & 16.81  \\
             Gemini Pro     & 5.46  & --    & --    \\
             PaLM 2 Unicorn & 5.88  & 2.94  & 22.27 \\
            \hline
            \multicolumn{4}{c}{\textbf{Dyck languages}} \\
             GPT-4-Turbo    & 6.99 & 28.46 & --     \\
             GPT-4          & 7.37 & 40.81 & 43.91  \\
             GPT-3.5-Turbo  & 1.28 & 6.05  & 2.08  \\
             Gemini Pro     & 2.25  & --    & --    \\
             PaLM 2 Unicorn & 0.38  & 6.43 & 17.77 \\
            \hline
        \end{tabular}
        \caption{Mistake finding accuracy for traces that contain mistakes (incorrect$_{mis}$).}
        \label{tab:split_scores_incorrect}
    \end{subtable}
    \caption{Mistake finding accuracy across 5 tasks for correct$_{mis}$ and incorrect$_{mis}$ traces. The combined scores of \autoref{tab:split_scores_correct} and \autoref{tab:split_scores_incorrect} make up \autoref{tab:benchmark_results}.}
\end{table*}

\end{document}